\documentclass[10pt,twocolumn,letterpaper]{article}

\usepackage{wacv}
\usepackage{times}
\usepackage{epsfig}
\usepackage{graphicx}
\usepackage{amsmath}
\usepackage{amssymb}
\usepackage{color}
\usepackage{overpic}
\usepackage{enumitem}
\usepackage{dblfloatfix}

\usepackage[pagebackref=true,breaklinks=true,letterpaper=true,colorlinks,bookmarks=false]{hyperref}

\wacvfinalcopy % *** Uncomment this line for the final submission

\def\wacvPaperID{599} % *** Enter the wacv Paper ID here

\ifwacvfinal\fi
\begin{document}

%%%%%%%%% TITLE
\title{ FreeLabel: A Publicly Available Annotation Tool based on Freehand Traces}

\author{Philipe A. Dias$^{1}$ \hspace{1.5cm} Zhou Shen$^{1}$ \hspace{1.5cm} Amy Tabb$^{2}$ \hspace{1.5cm}
Henry Medeiros$^{1}$ \\
$^{1}$Marquette University (EECE), USA \hspace{1.5cm} $^{2}$U.S. Department of Agriculture (USDA) \\ 
{\tt\small \{philipe.ambroziodias,zhou.shen,henry.medeiros\}@marquette.edu\hspace{1.5cm} amy.tabb@ars.usda.gov}
}

\maketitle
\ifwacvfinal\thispagestyle{empty}\fi

%%%%%%%%% ABSTRACT
\begin{abstract}
Large-scale annotation of image segmentation datasets is often prohibitively expensive, as it usually requires a huge number of worker hours to obtain high-quality results. Abundant and reliable data has been, however, crucial for the advances on image understanding tasks achieved by deep learning models. In this paper, we introduce FreeLabel, an intuitive open-source web interface that allows users to obtain high-quality segmentation masks with just a few freehand scribbles, in a matter of seconds. The efficacy of FreeLabel is quantitatively demonstrated by experimental results on the PASCAL dataset as well as on a dataset from the agricultural domain. Designed to benefit the computer vision community, FreeLabel can be used for both crowdsourced or private annotation and has a modular structure that can be easily adapted for any image dataset.\footnote{The citation information for this paper is: P.A. Dias, Z. Shen, A. Tabb, H. Medeiros, ``FreeLabel: a publicly available annotation tool based on freehand traces," in 2019 IEEE Winter Conference on Applications of Computer Vision (WACV). doi: 10.1109/WACV.2019.00010. This version is identical to the publisher's text, publisher's version avaliable \href{https://ieeexplore.ieee.org/document/8659167}{here}.}
\end{abstract}

%%%%%%%%% BODY TEXT
\section{Introduction}
The rapid rise in popularity of deep learning models in computer vision has brought a corresponding demand for labeled data. Depending on the image understanding task, the required annotations may range from tags at the image level (image classification), to bounding boxes (object detection) or pixel-level annotations (image segmentation).

Varied and high-quality image annotations are crucial for both training and evaluation of models that are accurate and robust. Currently, most of the Convolutional Neural Network (CNN) models successful at image understanding tasks \cite{chen_deeplab:_2018,he2018mask,Li2016FCIS} are pre-trained on the ImageNet \cite{deng2009imagenet} and COCO \cite{Lin2014coco} datasets, due to their large variability.

Manual labeling of large datasets is challenging and time-consuming. The costs reported for the COCO dataset in \cite{Lin2014coco} illustrate these difficulties. Containing over 2.5 million object instances, its labeling using Amazon's Mechanical Turk (AMT) required: $~20k$ worker hours for category labeling at image-level; $~10k$ hours for instance spotting; and, staggering, $~55k$ hours for instance segmentation.

To meet the need for large, labeled datasets, several approaches have been proposed. Different types of crowdsourcing have been used to generate labeled data quickly, from commercially available solutions such as the AMT, annotation parties \cite{everingham_pascal_2010}, volunteer/citizen science initiatives \cite{giuffrida_citizen_2018}, and custom-built pipelines \cite{chen_counting_2017}. 
\begin{figure}[h]
  \centering
  \includegraphics[width=.48\linewidth]{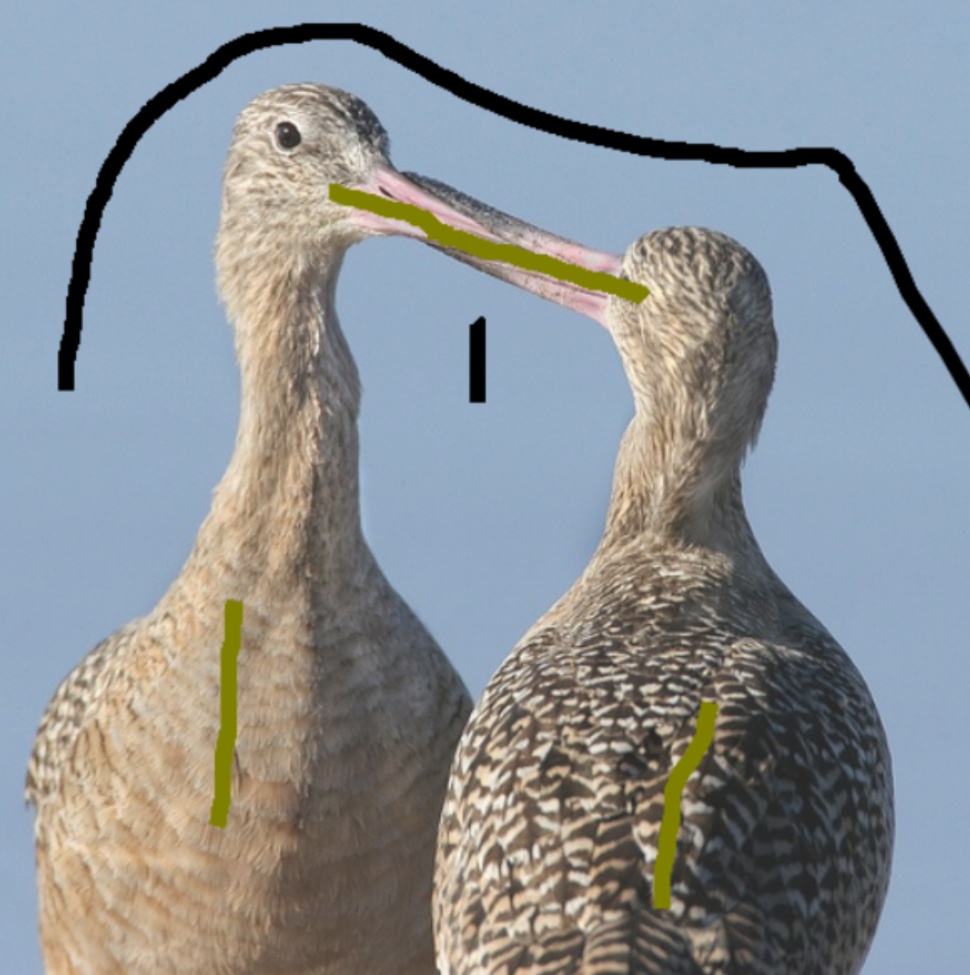}
  \includegraphics[width=.48\linewidth]{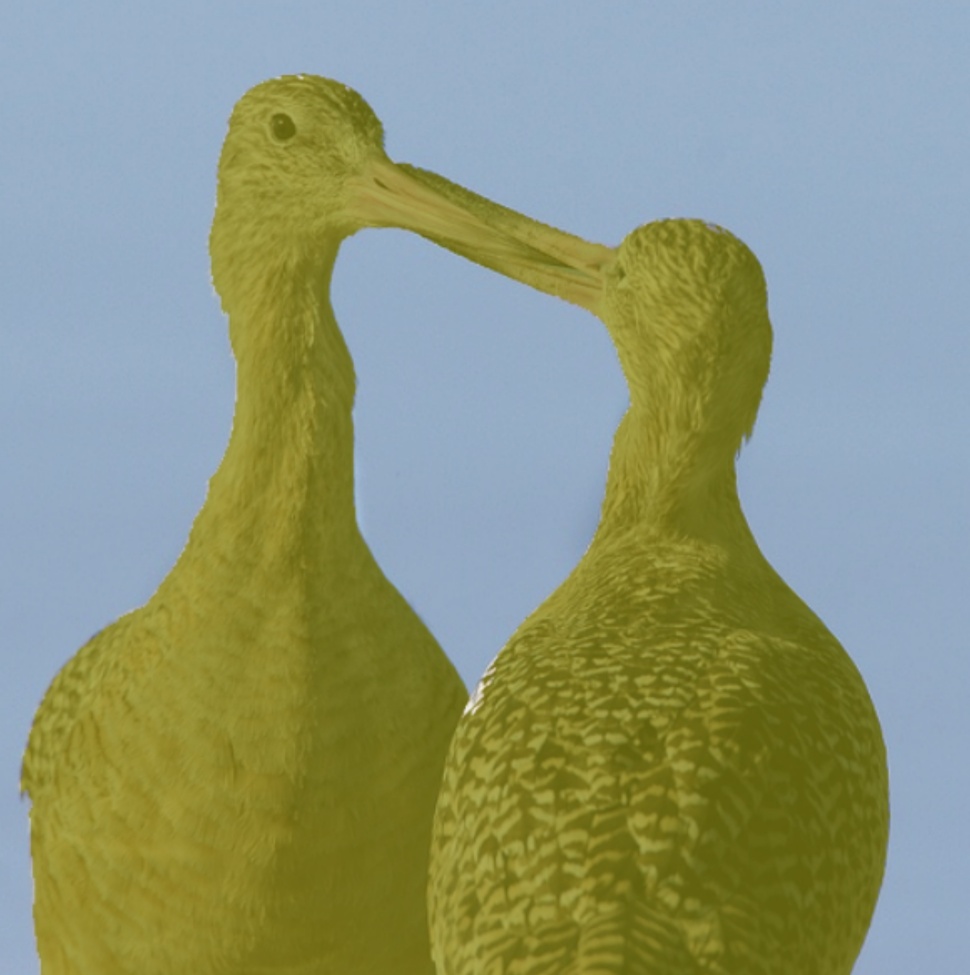}
    \caption{This paper describes an annotation tool that generates high-quality segmentation masks using simple freehand traces as input. From the few user traces illustrated in the left image, our FreeLabel tool outputs the object segmentation indicated by the yellow overlay in the right image.}
    \label{fig:ex1}
\end{figure}

Rather than selecting individual pixels, a popular strategy consists of approximating segmentations as polygons, which can be problematic for objects with complex boundary structures. Other strategies focus on labeling pre-segmented regions, such as superpixels \cite{caesar_coco-stuff:_2018,tangseng_looking_2017}. Although these strategies accelerate the annotation process, the segmentation quality is at risk in scenarios where the pre-computed regions fail to properly attach to boundaries.

To minimize the need for finely-annotated training data, the development of weakly-supervised training methods is also an active field of research. Strategies for the propagation of sparse annotations include graph cuts \cite{rother_grabcut:_2004}, level sets \cite{wang2014touchcut} and graphical models \cite{lin_scribblesup:_2016}. As the leaderboard of the PASCAL VOC 2012 dataset\footnote{http://host.robots.ox.ac.uk:8080/leaderboard/} shows, the performance of models trained in this way is still noticeably worse than models trained with fully annotated masks. 

We combine ideas from both the existing annotation tools and the field of semi-supervised learning to facilitate and minimize the amount of user interactions for annotating image segmentation masks, ultimately reducing labeling costs. Our contribution consists of a web-based tool, named FreeLabel, which allows the user to trace lines or ``freehand" scribbles of different thicknesses for the different categories present in an image (Figure \ref{fig:ex1}). These scribbles are propagated to the remaining unlabeled pixels using the Region Growing Refinement (RGR) algorithm introduced in \cite{dias_semantic_2018} for semantic segmentation refinement. Compared to other algorithms, RGR has the advantages of being fully unsupervised (thus category agnostic), simple to implement, with computational time and parameterization that allow quick and simple user interactions.

We assess the applicability of our tool in two contexts: the first is general object segmentation, exemplified by the PASCAL VOC dataset that has pixel-accurate labels for multiple different categories; and the second is the annotation of images of fruit tree flowers, which has applications in precision agriculture \cite{dias_apple_2018,dias_multispecies_2018}. In the first context, we analyze how long it takes for users to become familiar with our tool, and also the average annotation time and the segmentation quality they obtain in comparison with the official PASCAL ground-truth.

The first context serves as training for the second, where images of flowers of multiple fruit tree species are annotated \cite{dias_apple_2018}. In this scenario, we evaluate how well users can annotate images for which no ground-truth is available, and thus no intermediate feedback is provided.

%% Contributions
Our contributions to the state-of-the-art are:
\begin{enumerate}[noitemsep]
\item{a web-based tool, FreeLabel, for interactive annotation that is shown to be intuitive and effective, with users obtaining high-quality segmentations in an average time of $60$ seconds per object for the PASCAL dataset;}
\item{FreeLabel can be easily configured for any object category or dataset, an advantage inherited from the underlying unsupervised growing algorithm and the modular implementation of the tool;}
\item{public release of the tool at \url{coviss.org/freelabel}}
\item{the web-based structure of FreeLabel allows crowdsourcing and, when data privacy is of concern, private annotation using a local deployment.}
\end{enumerate}

\section{Related work}
\subsection{Segmentation datasets and labeling tools}
Introduced in 2005, the PASCAL VOC dataset \cite{everingham_pascal_2015} is the most widely-used dataset for visual object segmentation. Images within its 2012 \textit{valtrain} set contain a total of $6929$ segmented objects, distributed within $20$ different semantic categories. As reported in \cite{everingham_pascal_2010}, the process of annotating the images with pixel-level accuracy was extremely time-consuming, even though a $5$-pixel wide tolerance margin was allowed around each object.

The ImageNet dataset \cite{wang2014touchcut} with its $15$ million labeled images was crucial for the development of deep CNNs that revolutionized the state-of-the-art in image classification. Inspired by such success, the COCO dataset \cite{Lin2014coco} was introduced in 2015 to foster advances in object recognition, localization, and segmentation. It comprises $2.5$ million objects instances in $328k$ images, labeled by AMT workers using an adapted version of the OpenSurfaces interface \cite{bell2013opensurfaces}.

The OpenSurfaces interface resembles the LabelMe web-based annotation tool \cite{russell_labelme:_2008}, which was introduced in 2008 and is still widely used for segmentation annotation. Users provide object segmentations by tracing polygons along its boundary and typing the object name after completing the polygon. However, as mentioned in \cite{russell_labelme:_2008,Lin2014coco}, quality control is an important concern with this scheme. High-quality segmentations of objects with complex boundary structures require large numbers of vertices, leading to a trade-off between quality versus time spent to label each object. For annotation of the COCO dataset, its authors opted to minimize costs by collecting only one annotation for each instance, which required on average $79$ seconds per object. Yet, despite efforts such as quality verification steps, the dataset still contains some segmentation masks that poorly attach to the object boundaries \cite{dias_semantic_2018}.

The Cityscapes dataset for semantic urban scene understanding \cite{cordts2016cityscapes} was also annotated using layered polygons. To ensure that rich and high-quality pixel-level segmentation masks were obtained, its corresponding $5k$ images were annotated in-house. Over $1.5$h were required on average for annotation and quality control of each image with a restricted pool of high-quality annotators.

Alternative labeling strategies exploit superpixels to facilitate the annotation process. The interface used for labeling the COCO-Stuff dataset \cite{caesar_coco-stuff:_2018} combines SLICO superpixels \cite{achanta2012_slic} with a size-adjustable paintbrush tool that enables labeling of large regions at once. As mentioned by Tangseng et al. in \cite{tangseng_looking_2017}, superpixel errors can lead to significant annotation errors with this kind of interface. To minimize these artifacts, the authors described in \cite{tangseng_looking_2017} a interface that performs morphology-based boundary smoothing and allows the annotator to select the desired superpixel size to improve boundary adherence. Yet, this increases the complexity of the task, as the user has to try different configurations and label each superpixel individually.

Recently, an alternative approach for interactive segmentation was introduced in \cite{dextr2018}, where a CNN is trained to generate segmentation masks from extreme points specified by the user. The tool is shown to provide annotations of good quality in a timely manner, but requires supervised training and more computational resources.

\subsection{Weak and unsupervised segmentation}
{\bf Graph cuts.} Energy minimization approaches using the graph cuts paradigm are suited to interactive segmentation in that hard constraints are specified via squiggles for background and foreground classes \cite{boykov_graph_2006,boykov_interactive_2001,freedman_interactive_2005}. The popular GrabCut algorithm \cite{rother_grabcut:_2004} improved over interactive tools such as Intelligent Scissors (Magnetic Lasso) \cite{mortensen1995intelligent}, relaxing some of the labeling burden on the user. The user selects a bounding box of background pixels and can further edit the generated segmentation by drawing firm background/foreground traces. Gaussian Mixture Models (GMMs) are used for color modeling and a Gibbs energy is iteratively minimized using minimum cut. 

{\bf Level sets.} The level set approach has been used in segmentation since the 1990s, and can also be formulated as an energy minimization problem. Given an initialization, a boundary is evolved in the direction of a local minimum found via front propagation by solving partial differential equations \cite{cremers_review_2007}. An issue with level set implementations in the 2000s was runtime, and interactive approaches focused on reducing runtime using GPU implementation \cite{cates_gist:_2004,cremers_probabilistic_2007}. One approach allowed user input to adjust model parameters, in \cite{cates_gist:_2004}, while \cite{cremers_probabilistic_2007} reformulated energy functionals to incorporate user input. In \cite{liu_interactive_2012}, bounding-box initialization and the level set formalism were used for interactive segmentation. The TouchCut \cite{wang2014touchcut} interface exploits level-sets to grow segmentation masks from single points, which is effective when foreground and background colors are significantly different.

{\bf Propagation by pixel-affinity.} In a similar fashion that superpixel algorithms segment input images into clusters \cite{stutz2017sppx}, several matting and segmentation algorithms use low-level information such as texture, color affinity and spatial proximity to classify unlabeled regions based on sparse annotations \cite{chuang2001bayesian,chen2013knn,zhu2015targeting}. Similar methods have been used to refine segmentation masks predicted by CNNs \cite{krahenbuhl2012crf,chen2016dt,chen_deeplab:_2018}, as CNNs successfully exploit high-level context for semantic classification but fail to generate predictions with proper adherence to object boundaries. One such method is the {\it Region Growing Refinement} (RGR) \cite{dias_semantic_2018}, which combines Monte Carlo sampling of high-confidence samples with a region growing algorithm that is guided by spatial and color proximity between neighboring pixels. Selected as a building-block for FreeLabel, we describe more details of RGR in Section \ref{sec:methods}.

{\bf Joint propagation and CNN training.} Recent approaches aiming at interactive or weakly-supervised semantic segmentation focus on architectures in which the propagation of sparse annotations and the optimization of network parameters are performed jointly. Different works combine Fully Convolutional Networks (FCNs) with: GrabCut \cite{rajchl_deepcut:_2017}; superpixels and graphical modeling \cite{kolesnikov_seed_2016,lin_scribblesup:_2016}; novel loss functions and training strategies for weakly-supervised and interactive learning \cite{kolesnikov_seed_2016,tang_normalized_2018,mahadevan2018iter}. In \cite{aksoy2018semantic}, the idea of Laplacian matting matrices is combined with superpixels and a Deeplab-ResNet \cite{chen_deeplab:_2018} to identify layers (soft segments) that are semantically meaningful. For annotation of video sequences, in \cite{chen2018blazingly} a FCN is used to map input pixels onto an embedded space where pixels belonging to the same instance are close together, followed by a nearest-neighbor approach that classifies pixels based on reference masks provided at the first frame and on sparse user inputs.

\subsection{Good practices for design of annotation tools}
\label{sec:goodprac}
Vondrick et al. in \cite{vondrick_efficiently_2013} provide a set of best practices for crowdsourced video annotation, based on a three-year large scale study costing thousands of dollars for image annotation. A critical observation is that annotating platforms must aim at minimizing the cognitive load of the user. As backed by psychology studies \cite{schwartz2003paradox,bailey2006attention}, minimizing interruptions and choices help to reduce user anxiety and increase efficiency. Moreover, they observed that providing motivational feedback increases the workers' confidence that their work will not be rejected, which encourages workers to continue annotating.

Games With A Purpose (GWAP) exploit the idea that adding game-like elements to interfaces additionally motivates users to perform tasks of interest. The ESP Game \cite{von2004labeling} for image labeling is a widely known example: an image is shown to two players (users) and, without external communication, both enter possible words until a word is agreed upon. The common word becomes a label for the image. Other examples are the Peekaboom game for object localization \cite{von2006peekaboom}, Verbosity to collect commonsense facts about words \cite{von2006verbosity}, and Phylo for multiple sequence analysis \cite{kawrykow2012phylo}. 

Users play for the desire of being entertained, rather than for money or altruism \cite{von2008designing}. Timed response, score keeping, and randomness are important features for designing challenging and hence enjoyable games \cite{von2008designing}, as players are driven to play by the desire of increasing their skill level or to score higher than others. Compared to subjective and verbal instructions, scores are a more intuitive form of feedback to the user as they combine multiple aspects into a single performance metric. 
\begin{figure*}[t]
  \centering
  \includegraphics[width=.85\linewidth]{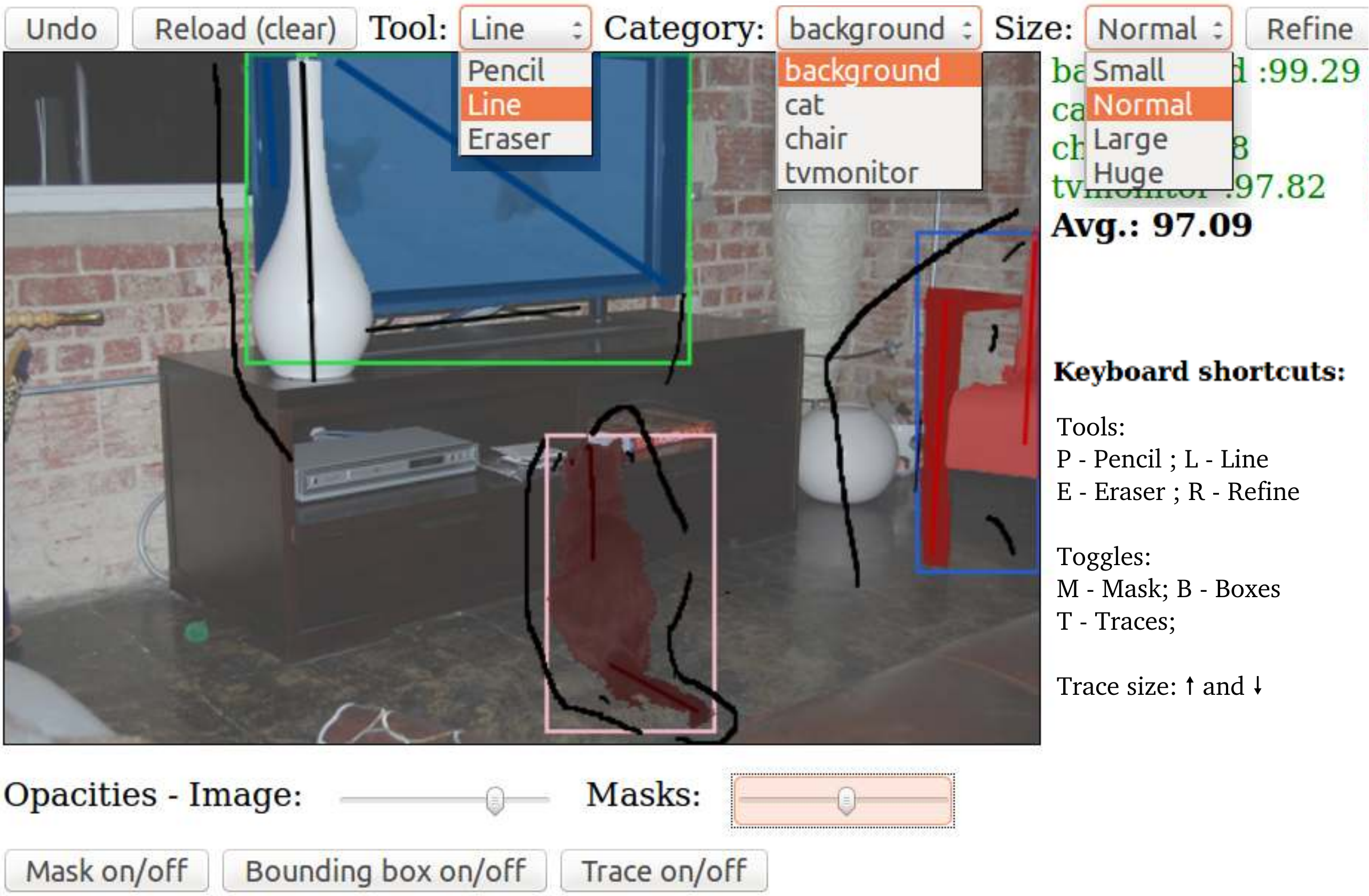}
    \caption{Diagram summarizing how the different modules of FreeLabel interact with each other. Users can draw with a freehand pencil or line segments. An eraser allows undoing small errors. Dialog boxes allow the user to select the object categories associated with the current trace, as well as adjust tool sizes. To help with visibility, other options such as opacity and masks are available via slider bars.}
    \label{fig:screen}
\end{figure*}

\section{Method}
\label{sec:methods}
Our objective is to develop a web-based labeling interface that: i) is intuitive to use; ii) allows users to quickly provide high-quality annotations; iii) can be easily adapted for different datasets and categories. 

As observed in Section \ref{sec:goodprac}, a good user interface should minimize the cognitive load on the user. Thus, instead of using propagation techniques that require supervised training or manual tuning of different sets of parameters, our tool exploits the RGR algorithm for unsupervised region growing. Based on related works, limitations of current tools and previous experiences with image annotation, we opted for designing a tool where the user input consists in simply drawing scribbles (freehand traces) or straight line segments on the images. 

By keeping all the parameterizations of the RGR algorithm fixed, we avoid any non-intuitive burden on the users. The quality of the segmentation provided by RGR is proportional to the amount and quality of initial seeds available. In this way, the user interaction to guide the growing process becomes quite intuitive, with simple guidelines: traces are grown based on color similarity and must be provided within the boundaries of the corresponding objects; thicker traces act as enforcement for the growing algorithm, since more seeds are available than for thin traces; if any region is incorrectly labeled by RGR, the user can easily correct it by adding a new trace of the correct category.

\begin{figure}[h]
  \centering
  \begin{overpic}[width=0.32\linewidth]{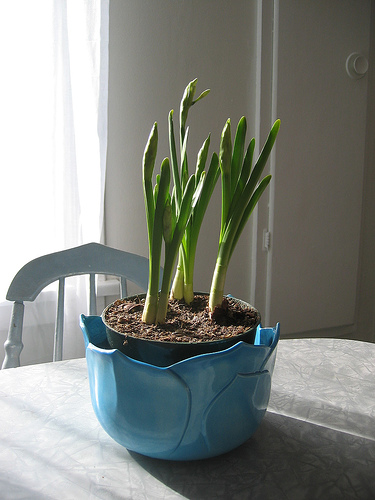}
  \put(0,0){\includegraphics[width=0.32\linewidth]{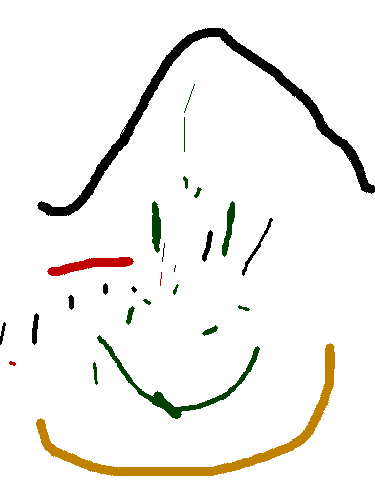}
       }
  \end{overpic}
  \includegraphics[width=0.32\linewidth]{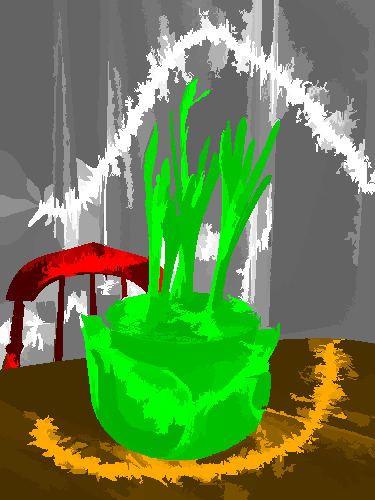}
  \begin{overpic}[width=0.32\linewidth]{images/imgPlant.jpg}
  \put(0,0){\includegraphics[width=0.32\linewidth]{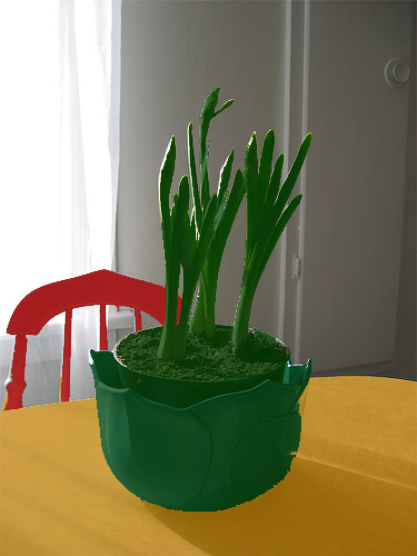}
       }
  \end{overpic}  \caption{Illustration of how traces are propagated to neighboring pixels. \textit{Left:} input traces drawn by the user. \textit{Center:} the brightness (intensity) of the color in each pixel is proportional to the score computed for its most likely category. For better visualization, \textit{background} traces are shown in black, while the \textit{background} likelihood is in grayscale from black (lowest) to white (highest). \textit{Right:} final segmentation obtained using maximum category likelihood per-pixel, with transparent \textit{background}.}
  \label{fig:grow}
\end{figure}

In addition to its simple formulation, we found the RGR implementation to be very suitable for multi-class segmentation annotation. Its growing process is class agnostic, propagating initial seeds into clusters regardless of seed label. This is advantageous in terms of running time, as the growing process has the same computational complexity regardless of the number of classes present in the image (average runtime lower than 1 second for PASCAL images \cite{dias_semantic_2018}). After clusters are formed for each set of seeds, they are classified into semantic categories by means of simple majority voting. Figure \ref{fig:grow} shows an example of this process, where each cluster is assigned to the class for which it contains the most labeled pixels. 

\subsection{FreeLabel Functionality}
\label{sec:funcs}
Figure \ref{fig:screen} shows a screenshot illustrating the functionality of our interface, together with an example of high-quality segmentation masks obtained from only a few user interactions. Three tools are available for drawing and adjusting traces using the mouse: 
\begin{itemize}[noitemsep]
\item \textit{Pencil \includegraphics[width=10pt]{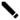}}: used for quickly tracing freehand scribbles. Once the user holds down the mouse's left-button, traces corresponding to the mouse trajectory are drawn. It is especially useful for regions that do not require high precision;
\item \textit{Line \includegraphics[width=10pt]{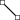}}: traces straight lines connecting the point where the user clicked the mouse button to the point where it was released. It is especially helpful for straight and thin structures, such as chairs' legs and animals' limbs.
\item \textit{Eraser  \includegraphics[width=10pt]{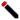}}: used to correct imprecisions in provided scribbles, such as small portions protruding outside the corresponding object's boundary.
\end{itemize}

Each tool can be configured with four different thicknesses: \textit{small} ($1$px thick), \textit{normal} ($2$px), \textit{large} ($4$px) or \textit{huge} ($8$px). After tracing scribbles over the image, the user can invoke the RGR algorithm by simply clicking the \textit{Refine} button, which automatically grows segmentation masks from the provided traces. To annotate smaller objects, the user can zoom in/out using the mouse scroll, as in any modern web-browser. Finally, keyboard shortcuts are available for all the commands to facilitate the annotation process.

In addition to intuitive commands, visualization is another key factor that impacts the labeling experience and annotation quality. Similar to the PASCAL, COCO, and other datasets, a specific color is associated to the traces and masks of each category. For the background, traces are shown in black and the masks are invisible. To handle scenarios where the image is too dark or contains colors with poor contrast to traces and/or masks, our interface allows the user to control the brightness (opacity) of both the image and the segmentation masks using the sliders under the canvas. Moreover, masks and traces can be hidden/shown with the click of the corresponding toggle buttons. 

\subsection{Implementation}
\begin{figure}[h]
  \centering
  \includegraphics[width=\linewidth]{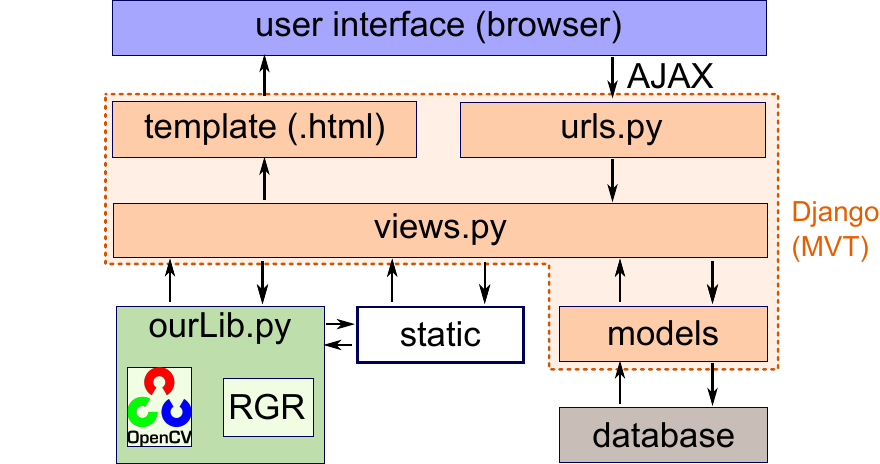}
    \caption{Diagram summarizing how the different modules of FreeLabel interact with one another.}
    \label{fig:flow}
\end{figure}

Our FreeLabel tool for segmentation annotation relies on three main building blocks: a graphical user interface (GUI), the Django framework, and the RGR algorithm. Figure \ref{fig:flow} summarizes the relationships.

An important criterion for our design choices concerned how easy the user's inputs and the RGR algorithm could be combined for the computation of segmentation masks. Aiming at an open-source web interface, we adapted RGR's original MATLAB implementation to Python and opted for the Django platform as the web framework.

Django \cite{django2018} is a free, open source Python framework that follows the Model-View-Template architectural pattern. The \textit{Model} layer allows access to database information without requiring any knowledge of the intricacies of database rules. The \textit{View} logic layer of Django handles the communication between the \textit{Model} and the \textit{Templates}, which correspond to the exhibition layers that define what is shown to users through the browser. 

Using Figure \ref{fig:flow} as guidance, a top-down walk-through of our tool's implementation starts with the graphical interface displayed by the web browser to the user. The design and functionality described in Section \ref{sec:funcs} and exemplified in Figure \ref{fig:screen} are implemented as customized Django templates, using HTML/Javascript. For actions requiring the execution of Python commands, the template (.html) file will trigger an AJAX call that is mapped to a corresponding function in \textit{views} (.py). This layer mediates the access to the database (through the \textit{Model} layer), static files or any customized Python function.

Aiming at a modular implementation that can be easily tailored for different datasets or configurations, we package the implementation of RGR and other custom functions into a separate Python library (\textit{ourLib.py}). This includes functions using the OpenCV \cite{opencv_library} library, which are responsible for image loading and converting the outputs of RGR from mathematical arrays to images for visualization.

\setcounter{figure}{5}
\begin{figure*}[b]
  \centering
  \includegraphics[width=.24\linewidth]{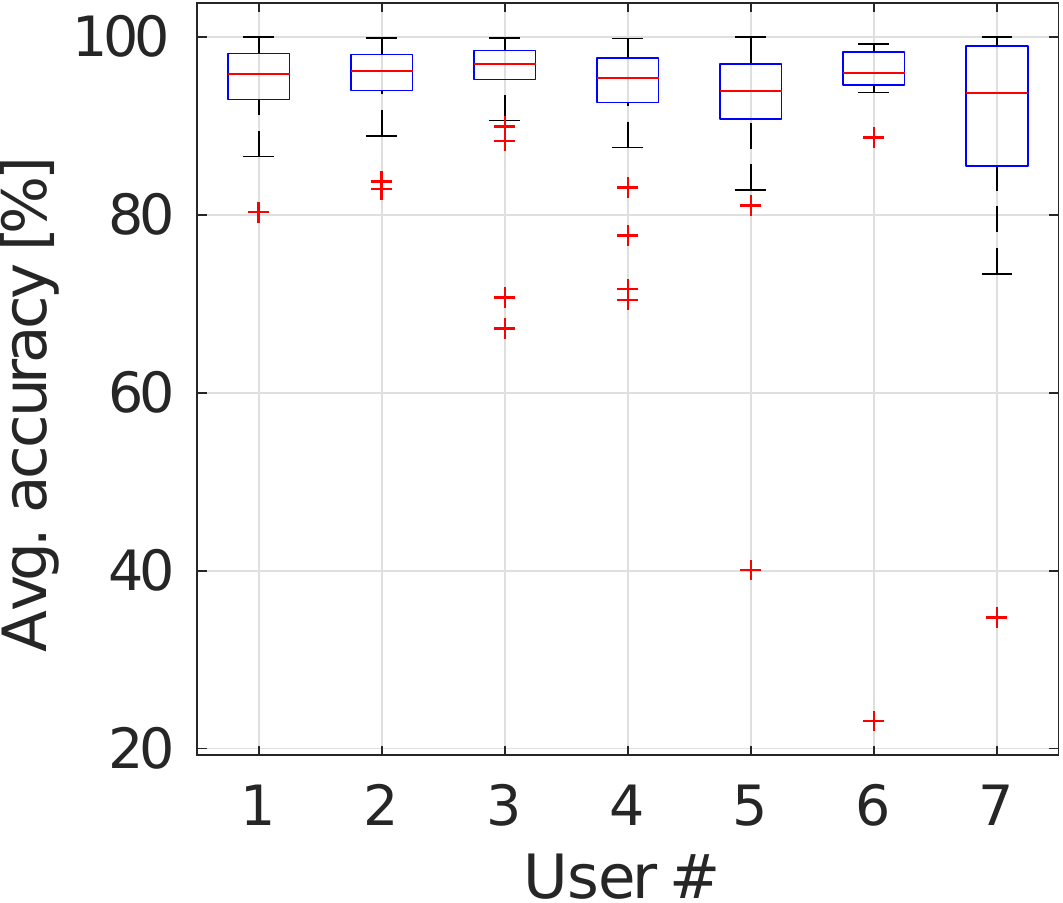}\hspace{0.008\linewidth}
  \includegraphics[width=.24\linewidth]{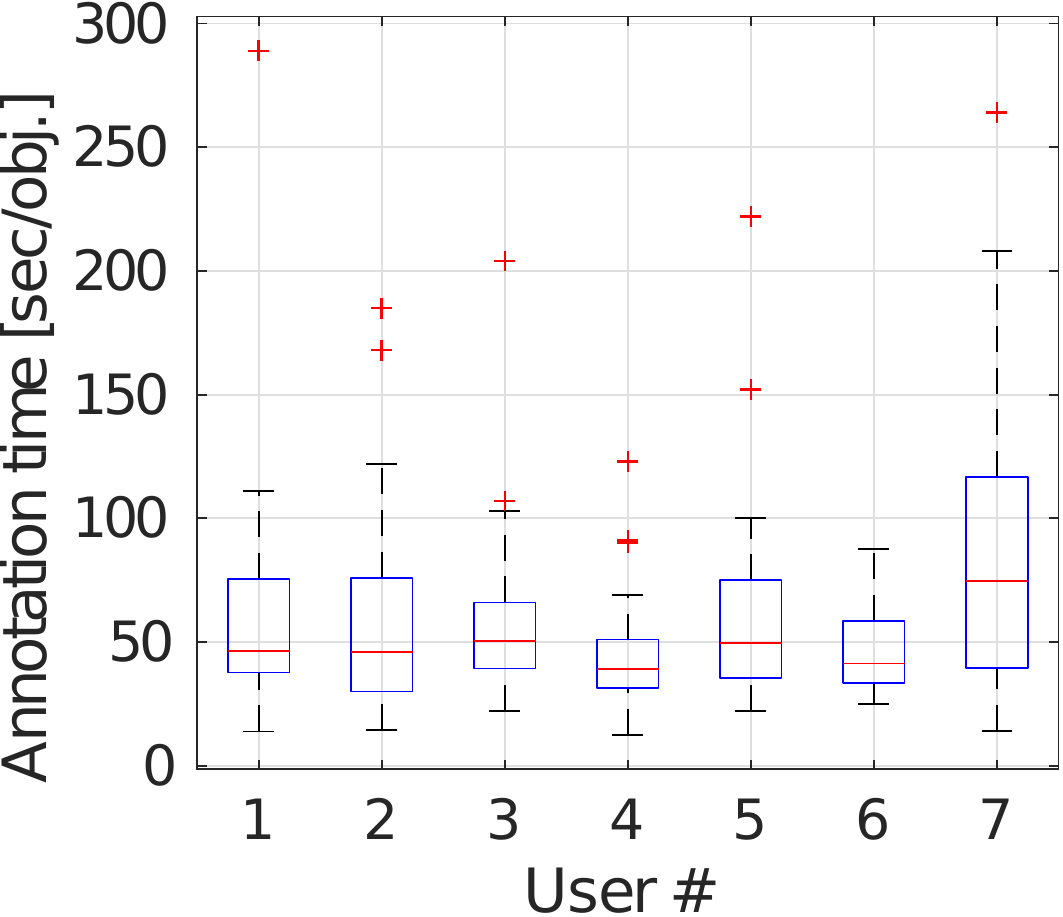}\hspace{0.008\linewidth}
  \includegraphics[width=.24\linewidth]{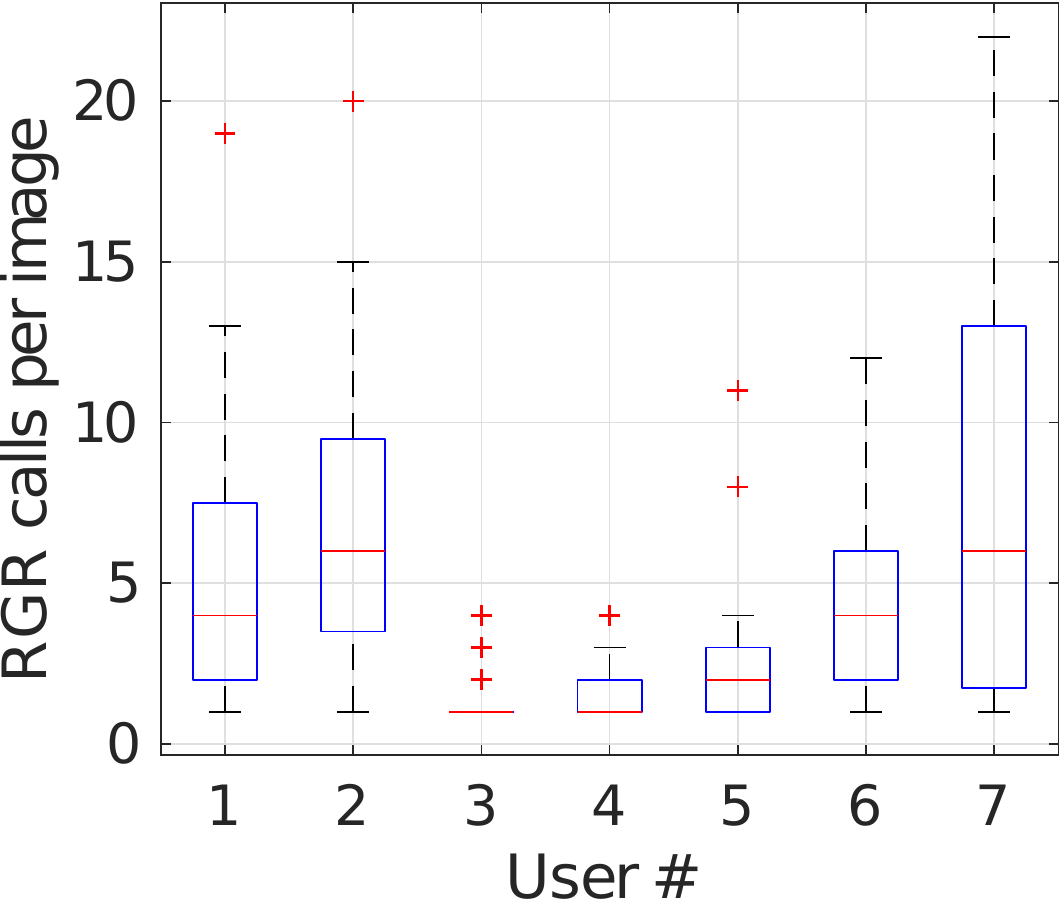}\hspace{0.008\linewidth}
  \includegraphics[width=.24\linewidth]{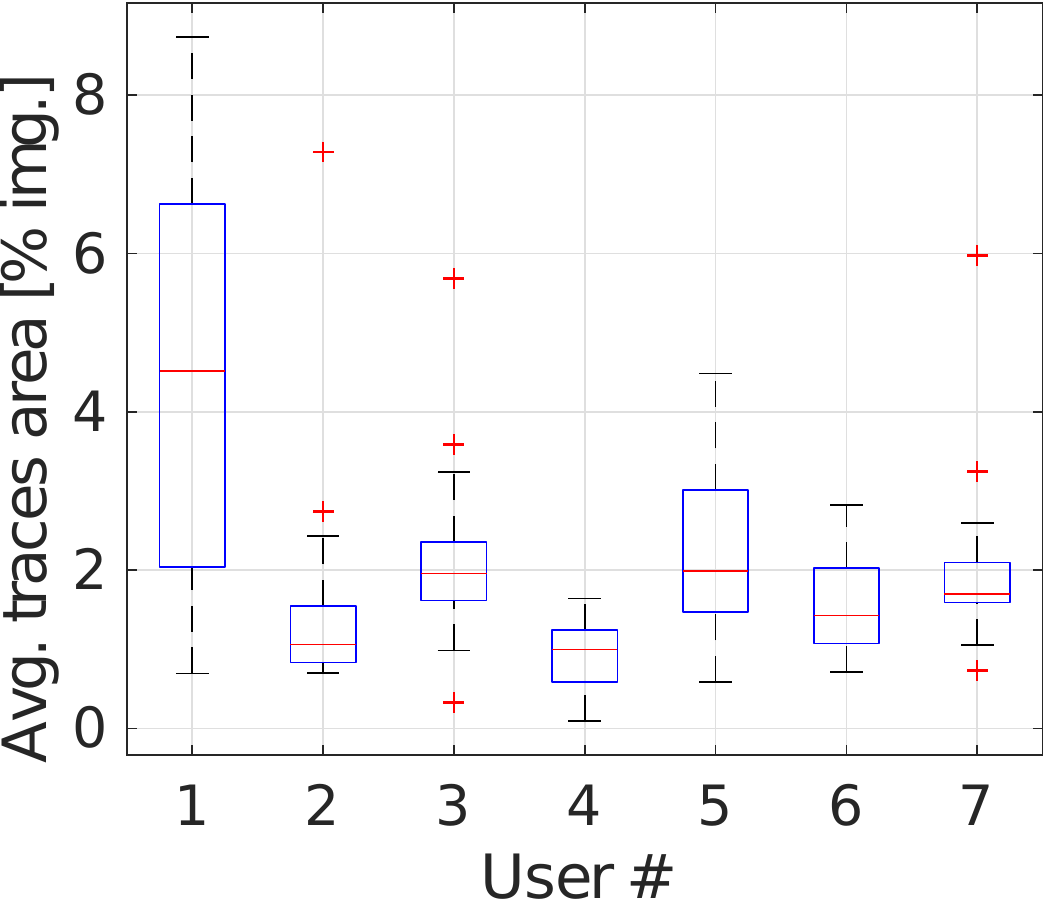}  
  \caption{Distribution of the obtained accuracies, annotation times, number of \textit{Refine} calls and average image area covered by user traces for annotating images from the PASCAL dataset.}
  \label{fig:statsUsr}
\end{figure*}

RGR is used as the core component of FreeLabel, and adapted in two minor aspects to compose the annotation tool. The original algorithm described in \cite{dias_semantic_2018} focuses on the refinement of a CNN's semantic segmentation predictions, a scenario with coarse segmentation masks as input. While for that case sampling fewer seeds is beneficial to filter out false-positives, in our scenario we aim at minimizing the required number of user interactions. Since the user inputs tend to be sparse but highly-accurate, we increase the percentage of seeds sampled in each Monte Carlo iteration to $75\%$, with $8$ iterations per run. Moreover, we remove RGR's constraint that automatically classifies as background any pixel significantly distant from labeled neighbors in terms of appearance and spatial position. By removing this constraint, RGR will assign to each unlabeled pixel the category provided for its nearest neighbor, regardless of how far they might be. If the propagated label is incorrect, the user can easily improve the segmentation by tracing an additional scribble to the corresponding region.

\section{Experiments and Results}
We evaluate our tool in terms of: i) quality of the obtained segmentation masks; and ii) time required by users to annotate images using FreeLabel. To that end, we defined first a task where users were asked to annotate images from the PASCAL VOC 2012 dataset. We opted for this dataset as it contains good quality segmentations of multiple object categories and is widely used by the computer vision community, such that it represents a good reference standard for anyone searching for a suitable annotation tool. 

Inspired by the idea of GWAP, we designed a game-like version of FreeLabel for the annotation of PASCAL images. Ideally, users must provide high-quality segmentation but also be as quick as possible, which represents a trade-off for which it is difficult to provide the annotators with clear guidelines. We therefore employ a game with a simple unified score metric that combines annotation time and mean average precision ($mAP$) between the obtained masks and corresponding ground-truth annotations, which is computed according to the official PASCAL metrics. The quality of the segmentation must be the main priority, while the time spent on each image is a secondary concern. Thus, as summarized in Figure \ref{fig:scores}, we use accuracy ($mAP$) as the base factor for score computation, with a ``bonus'' multiplying factor that is proportional to the time spent on each image. 

\setcounter{figure}{4}
\begin{figure}[h]
  \centering
  \includegraphics[width=0.9\linewidth]{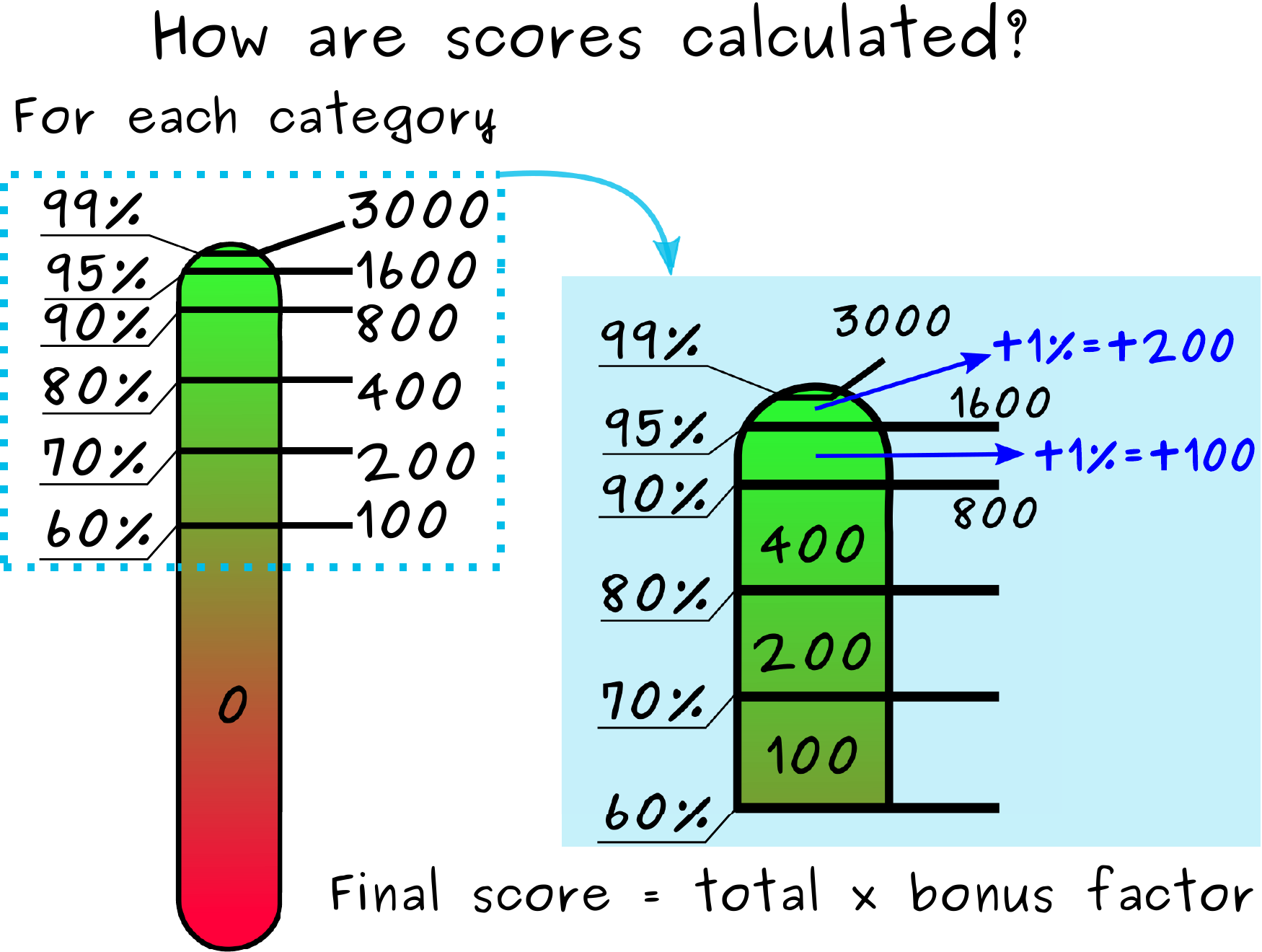}
  \caption{Score chart presented as reference for the game where users are asked to label PASCAL images in an accurate and timely manner.}
  \label{fig:scores}
\end{figure}

The main goal of this metric is to constitute feedback that tells the user how well he/she is performing the task, such that we do not focus on a more rigorous formulation for score computation. Instead, we aim at motivating the user to obtain the highest accuracy as possible by increasing the base score progressively as the $mAP$ approaches $100\%$. 

Let $N$ denote the number of objects in an image. Based on the performance of expert labelers, we roughly estimated an expected time of $60$ seconds for an image with $N=1$, plus an extra $30$ seconds per object when $N\geq2$. To motivate users to be quick, we thus multiply the base score with a bonus factor according to Eq.\ref{eq:bonus}: $2\times$ if the user annotates the image in the expected time $T$, linearly decaying to $1\times$ if the annotation time $t$ takes longer than $2T$.  
\begin{align}
\label{eq:bonus}
bonus &= \max(2+\frac{T-t}{T},1) \\
T &= 60+30\times(N-1) [sec]. \nonumber
\end{align}

After showing the participants a training video, we asked seven different users to label an average of $25$ images each, in a task expected to take $1$ hour. We followed the official PASCAL annotation guidelines \cite{everingham_pascal_2010}, indicating with bounding boxes the objects to be annotated by the users.

Figure \ref{fig:statsUsr} summarizes the average accuracy ($mAP$) and average time needed to annotate the different objects in the images. Overall, users provided segmentations with $92.8\%$ overlap with the ground-truth masks, at a mean pace of $61.3$ seconds per object. As a reference, this is significantly quicker than the average $79$ sec/object required for annotating the COCO dataset using the OpenSurfaces tool \cite{Lin2014coco}.

\setcounter{figure}{6}
\begin{figure*}[t]
  \centering
	\includegraphics[width=0.16\linewidth]{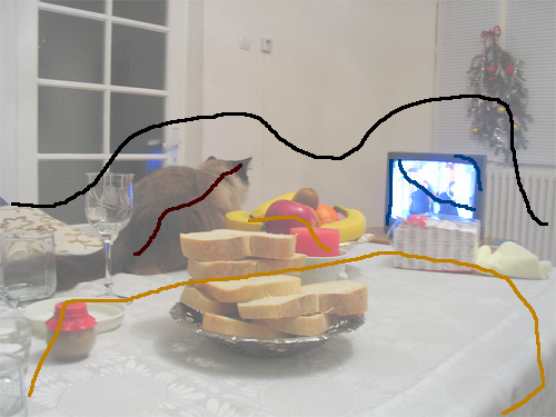}  
    \includegraphics[width=0.16\linewidth]{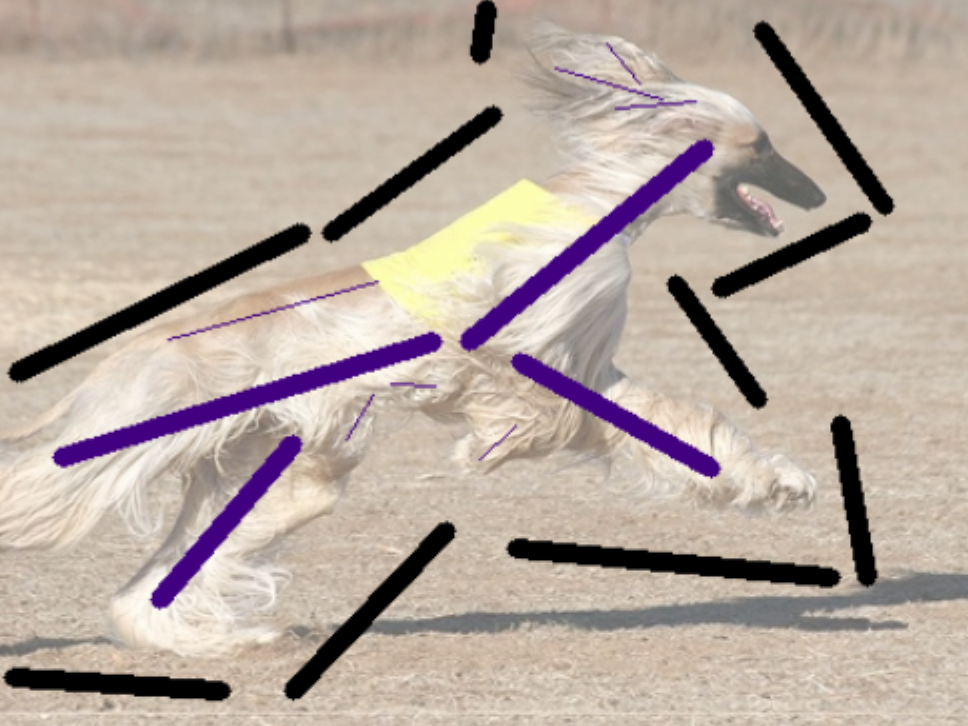}  
    \includegraphics[width=0.16\linewidth]{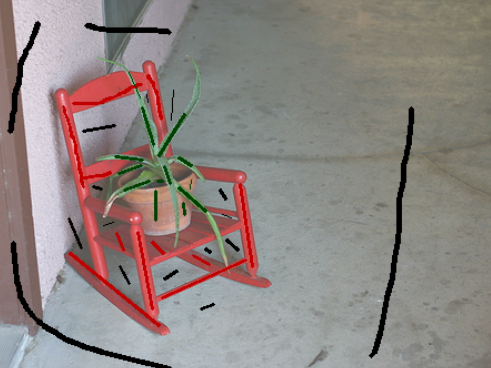}  
    \includegraphics[width=0.16\linewidth]{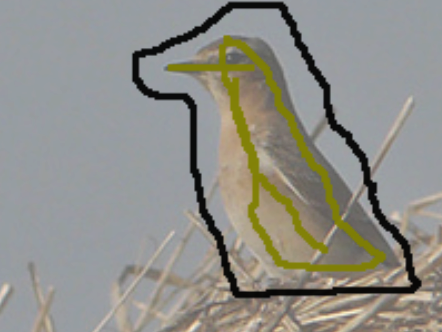}  
    \includegraphics[width=0.16\linewidth]{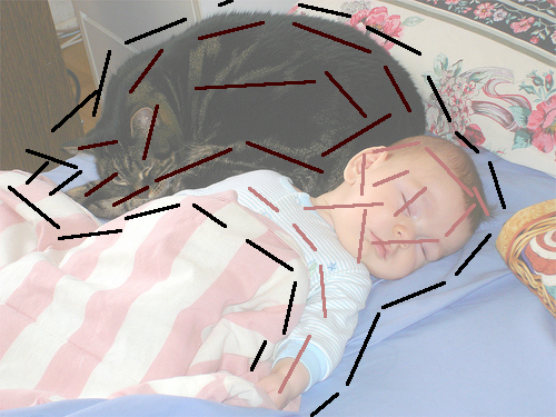}  
    \includegraphics[width=0.16\linewidth]{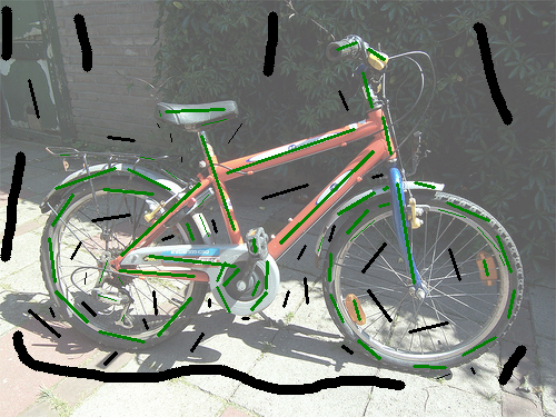}  	   
  	\includegraphics[width=0.16\linewidth]{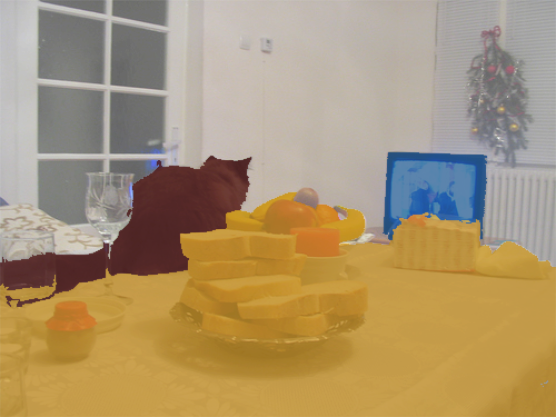}   
  	\includegraphics[width=0.16\linewidth]{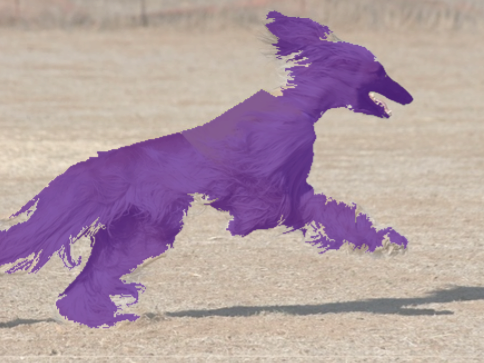}
\includegraphics[width=0.16\linewidth]{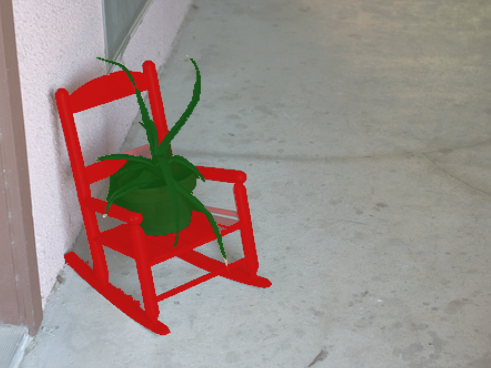}  
\includegraphics[width=0.16\linewidth]{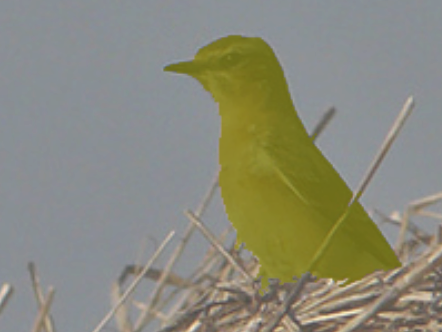}  
\includegraphics[width=0.16\linewidth]{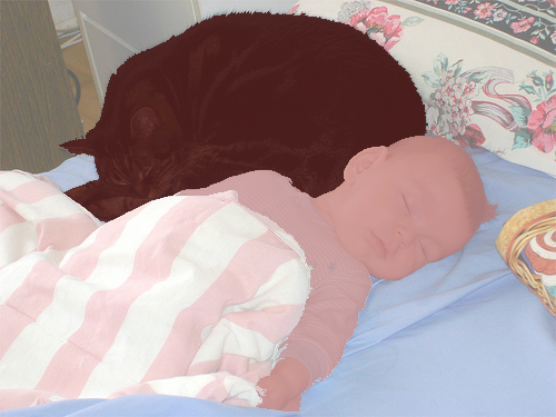} 
\includegraphics[width=0.16\linewidth]{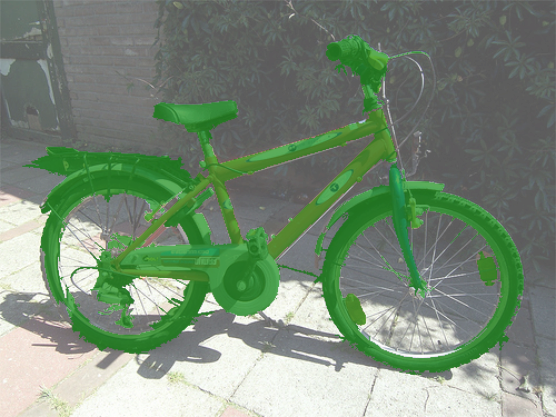}    	
\caption{Examples of annotations provided by users for the PASCAL dataset using FreeLabel. \textit{Top:} user annotations. \textit{Bottom:} final grown mask generated by FreeLabel from the corresponding inputs.}
  \label{fig:mosaic}
\end{figure*}S

We also observed which strategies were adopted by the most successful users. The two right-most plots in Figure \ref{fig:statsUsr} summarize the frequency of usage of the \textit{Refine} button by each user and the average image area covered by their scribbles, respectively. User $\#2$ exemplifies the usefulness of interactivity using RGR: by frequently using the \textit{Refine} option, this user obtained one of the highest accuracy averages, with fewer low-quality outliers. This user also drew fewer traces and thus finished the task faster than others who provided annotations of similar quality.

\begin{figure}[!h]
  \centering
  \includegraphics[width=0.92\linewidth]{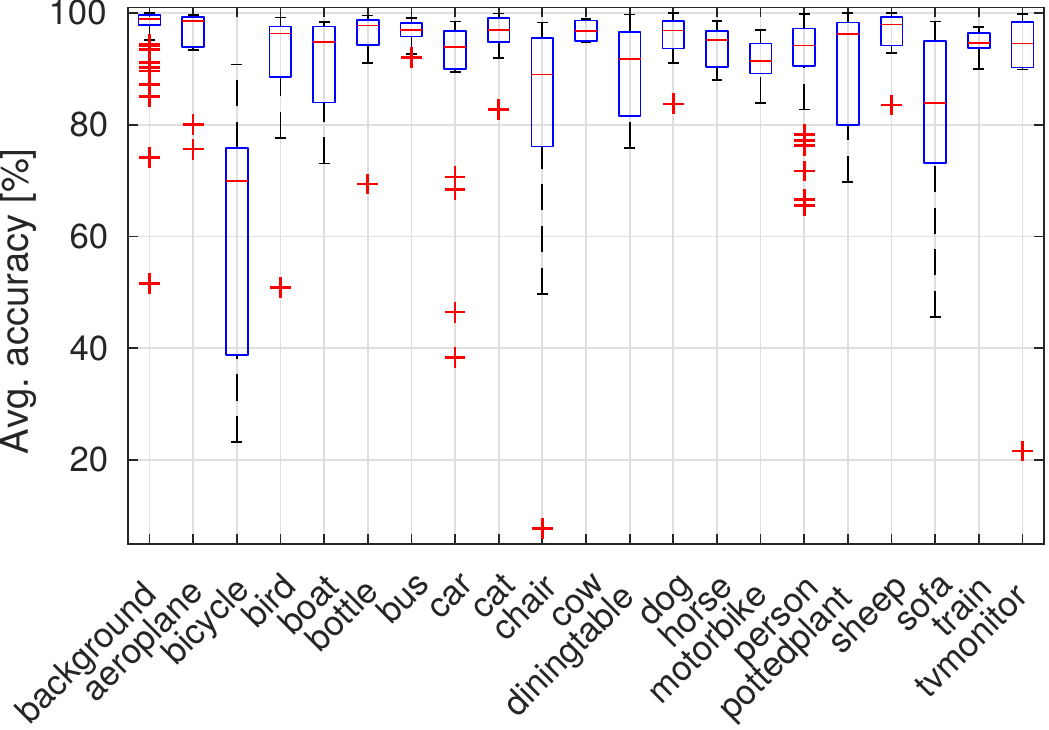}
  \caption{Distribution of average accuracy for objects of the different categories in the PASCAL dataset.}
  \label{fig:statsCat}
\end{figure}

Figure \ref{fig:statsCat} allows an analysis per object category that further highlights the qualities of FreeLabel. As the median values of $95.5\%$ overall accuracy and $50.1$ seconds per object suggest, the presence of outliers is confirmed by inspecting results for categories such as \textit{bicycle, chair} and \textit{pottedplant}. These are notably harder to label than instances from classes like \textit{airplane, cows} and \textit{trains}, which present fewer enclosed regions or thin structures. However, despite requiring longer annotation times, high-quality segmentations can still be obtained for such harder categories. Figure \ref{fig:mosaic} is a compilation of annotation examples provided by the users, with the \textit{bicycle} example illustrating the quality of segmentation that can be obtained even for harder cases.
\begin{figure*}[t]
	\centering
    \includegraphics[width=0.32\linewidth]{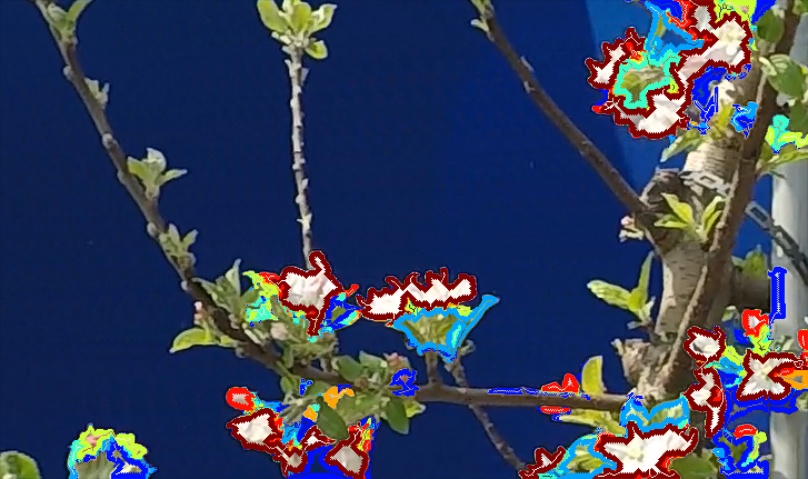}
    \includegraphics[width=0.32\linewidth]{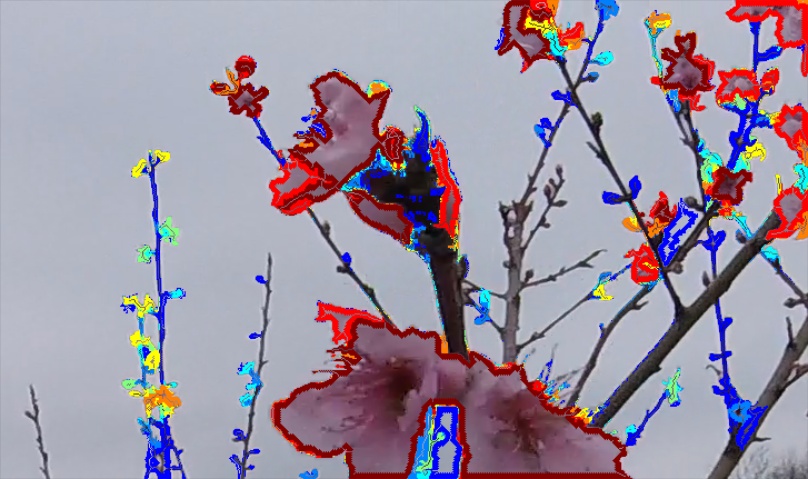}
    \includegraphics[width=0.32\linewidth]{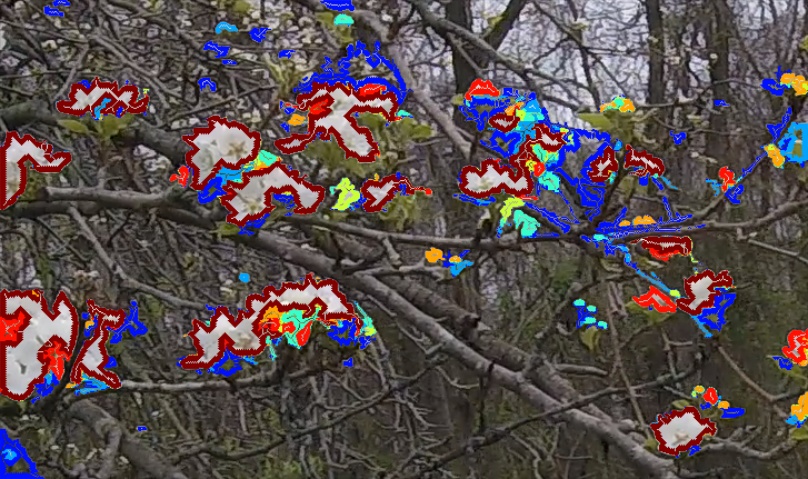}
  	\caption{Examples of flower annotations provided by users using FreeLabel. The colormap boundaries illustrate how many users labeled the enclosed regions as flower. Colors proportionally range from dark blue (one user) to dark red (all users labeled it as flower).}
  	\label{fig:mosaicFl}
\end{figure*}

\subsection{Annotation of unlabeled images}
To demonstrate the suitability of FreeLabel for the realistic scenario of annotating unlabeled datasets, we performed experiments where $8$ users were asked to annotate images of a significantly different dataset. We chose the dataset made publicly available in \cite{Tabb2018Datasets,dias_multispecies_2018}, which contains images of multiple species of fruit-flowers that were acquired under varied conditions. Since these are high-resolution images ($2704\times1520$px) containing dozens of small flowers, we decided to split each image into $16$ blocks of $676\times380$px.

With the lessons learned from the PASCAL experiments, we designed a new training sequence (video available together with the tool) that emphasizes good strategies for efficient labeling with FreeLabel. Before annotating the flowers, all users were required to annotate $10$ PASCAL images with a minimum accuracy of $90\%$ per category. Our rationale is that annotating the PASCAL images in a game-format works as a training session in which the users become familiar with the interface and grasp the main guidelines for annotating any type of image segmentation dataset.

Preliminary experiments indicated that the lack of performance feedback harms the motivation of the users and, as consequence, the quality of the segmentations obtained. Hence, we structured the annotation sessions such that each user was required to label $9$ blocks of different flower images, in batches of $3$ blocks each. Each batch contained $2$ non-annotated blocks and $1$ block for which ground-truth was available. We used the ground-truth image blocks as checkpoints: if the segmentation provided by the user did not meet a certain accuracy threshold, the user would have to redo the entire batch of $3$ images. The ground-truth annotations are never shown to the users, such that while only every third image is actually used to compute the average accuracy, we ``deceive'' the users to believe that all images are verified and must thus be accurately labeled. Moreover, we used a rather lower accuracy threshold of $70\%$, as the main intent is just to avoid very poor annotations.

Results demonstrate the effectiveness of this strategy for the annotation of unlabeled images. In Figure \ref{fig:mosaicFl}, the colormap progressively ranging from blue to red illustrates for each enclosed region how many users labeled it as flower. This representation qualitatively demonstrates how the annotations provided by the different users for the three different datasets converge to ideal segmentation masks. Such convergence suggests that majority voting can be used to approximate ideal segmentation masks, which we then use to statistically evaluate the variability of the annotations provided for images without ground-truth. 
\begin{figure}[h]
	\centering
    \includegraphics[width=0.98\linewidth]{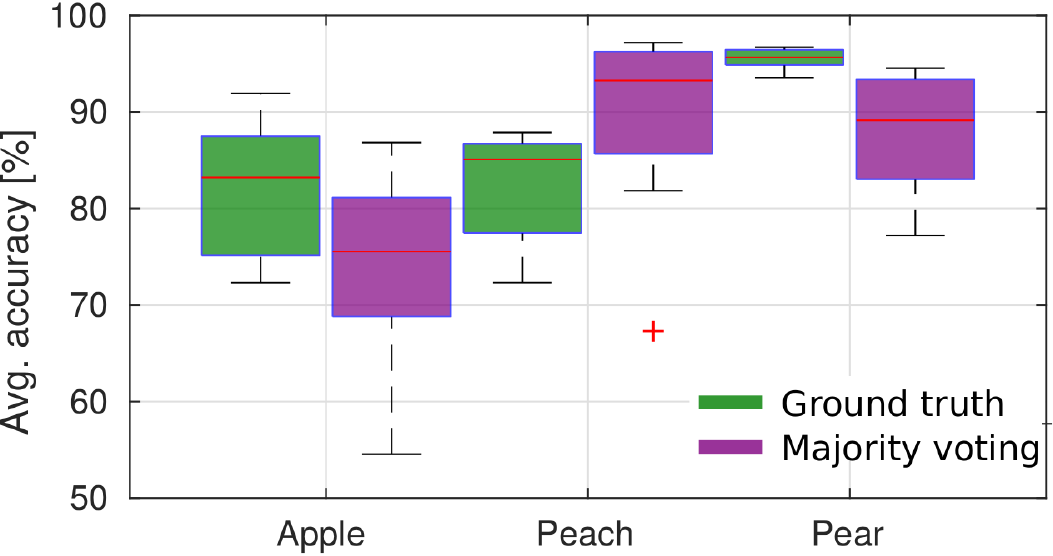}
  	\caption{Distribution of the average accuracy obtained by the users for annotation of flower datasets.}
  	\label{fig:statsflowers}
\end{figure}

Figure \ref{fig:statsflowers} summarizes the average accuracy and deviations observed for the images with and without ground-truth available (in green and purple, respectively). The average overlaps between the segmentations provided by the users and the available ground-truth masks were higher than $80\%$ for the three different datasets, reaching $95.5\%$ for the \textit{Pear} image. The higher deviations for the \textit{Apple} and \textit{Peach} datasets are mostly associated with the annotation of small flower buds and mistakes related to bright leaves on the apple images. Such mistakes are visible as well in the examples in Figure \ref{fig:mosaicFl}. Finally, the deviations observed for ground-truth images are similar to the ones observed for the images without ground-truth, which indicates a somewhat consistent performance of users for both groups of images.

\section{Conclusion}
We introduced FreeLabel, an interactive interface for fast and high-quality annotation of image segmentation datasets. In contrast to annotation tools that require drawing polygons fully enclosing objects to be segmented, FreeLabel simplifies the user interactions to freehand scribbles and straight lines. By means of the unsupervised algorithm known as RGR, such inputs are grown into segmentation masks that tightly adhere to actual object boundaries.

FreeLabel has a modular design and relies solely on open-source libraries, as we aim at a publicly available tool that can be easily adapted for annotation of a wide range of datasets. Its web-based arrangement can be deployed both locally or in external servers, allowing annotations through both private (confidential) or crowdsourced strategies.

Our experiments demonstrate that segmentations with high overlap to ground-truth annotations of the PASCAL dataset can be obtained in a matter of seconds. Through short tutorial videos and a game-like version of FreeLabel, users quickly learned how to use the tool and were capable of properly annotating significantly different datasets.

As future work, we intend to accelerate the RGR algorithm and evolve FreeLabel into an interactive tool that automatically grows the user scribbles in real-time. With minor adjustments, we believe FreeLabel could be also efficiently used with tablets and mobile devices. We also consider combining majority voting and GWAP for annotation of unlabeled datasets, exploiting cooperative and antagonistic roles for user motivation and annotation quality control.

Finally, we plan to hire AMT workers for larger scale image annotation using FreeLabel. Feedback received from $5$ AMT workers hired as a preliminary experiment included encouraging comments such as \textit{``I was surprised how well the bounding tools worked. They seemed to accurately pick up my responses''}, and \textit{ ``the interface was easy to understand for anyone mildly familiar with MS paint''}.

{\bf Acknowledgement. } This work was supported by USDA ARS agreement \#584080-5-020. Mention of trade names or commercial products in this publication is solely for the purpose of providing specific information and does not imply recommendation or endorsement by the U.S. Department of Agriculture. USDA is an equal opportunity provider and employer. We thank our colleagues Abubakar Siddique, Enrico Prampolini, Reza Mozhdehi, Jamir Jyoti, Brian Stumph, Scott Wolford, and Larry Crim who collaborated with data collection. We thank NVIDIA for providing the GPU used in this work.
{\small
\bibliographystyle{ieee}
\bibliography{refs}

\begin{thebibliography}{10}\itemsep=-1pt

\bibitem{django2018}
{Django} (version 2.0).
\newblock \url{https://djangoproject.com/}.
\newblock Accessed: 2018-09-13.

\bibitem{achanta2012_slic}
R.~Achanta, A.~Shaji, K.~Smith, A.~Lucchi, P.~Fua, and S.~S{\"{u}}sstrunk.
\newblock {SLIC superpixels compared to state-of-the-art superpixel methods}.
\newblock {\em IEEE Transactions on Pattern Analysis and Machine Intelligence},
  34(11):2274--2281, 2012.

\bibitem{aksoy2018semantic}
Y.~Aksoy, T.-H. Oh, S.~Paris, M.~Pollefeys, and W.~Matusik.
\newblock Semantic soft segmentation.
\newblock {\em ACM Transactions on Graphics}, 37(4):72, 2018.

\bibitem{bailey2006attention}
B.~P. Bailey and J.~A. Konstan.
\newblock On the need for attention-aware systems: Measuring effects of
  interruption on task performance, error rate, and affective state.
\newblock {\em Computers in Human Behavior}, 22(4):685 -- 708, 2006.
\newblock Attention aware systems.

\bibitem{bell2013opensurfaces}
S.~Bell, P.~Upchurch, N.~Snavely, and K.~Bala.
\newblock {OpenSurfaces: A richly annotated catalog of surface appearance}.
\newblock {\em ACM Transactions on Graphics}, 32(4):111, 2013.

\bibitem{boykov_graph_2006}
Y.~Boykov and G.~Funka-Lea.
\newblock Graph {Cuts} and {Efficient} {N}-{D} {Image} {Segmentation}.
\newblock {\em International Journal of Computer Vision}, 70(2):109--131, Nov.
  2006.

\bibitem{boykov_interactive_2001}
Y.~Y. Boykov and M.~P. Jolly.
\newblock Interactive graph cuts for optimal boundary amp; region segmentation
  of objects in {N}-{D} images.
\newblock In {\em {IEEE} {International} {Conference} on {Computer} {Vision}},
  volume~1, pages 105--112 vol.1, 2001.

\bibitem{opencv_library}
G.~Bradski.
\newblock {The OpenCV Library}.
\newblock {\em Dr. Dobb's Journal of Software Tools}, 2000.

\bibitem{caesar_coco-stuff:_2018}
H.~Caesar, J.~Uijlings, and V.~Ferrari.
\newblock {COCO}-{Stuff}: {Thing} and {Stuff} {Classes} in {Context}.
\newblock In {\em {IEEE} {Conference} on {Computer} {Vision} and {Pattern}
  {Recognition}}, page~10, 2018.

\bibitem{cates_gist:_2004}
J.~Cates, A.~Lefohn, and R.~Whitaker.
\newblock {GIST}: an interactive, {GPU}-based level set segmentation tool for
  3d medical images.
\newblock {\em Medical Image Analysis}, 8(3):217--231, Sept. 2004.

\bibitem{chen2016dt}
L.-C. Chen, J.~T. Barron, G.~Papandreou, K.~Murphy, and A.~L. Yuille.
\newblock Semantic image segmentation with task-specific edge detection using
  {CNNs} and a discriminatively trained domain transform.
\newblock In {\em IEEE Conference on Computer Vision and Pattern Recognition},
  pages 4545--4554, 2016.

\bibitem{chen_deeplab:_2018}
L.~C. Chen, G.~Papandreou, I.~Kokkinos, K.~Murphy, and A.~L. Yuille.
\newblock {DeepLab}: {Semantic} {Image} {Segmentation} with {Deep}
  {Convolutional} {Nets}, {Atrous} {Convolution}, and {Fully} {Connected}
  {CRFs}.
\newblock {\em IEEE Transactions on Pattern Analysis and Machine Intelligence},
  40(4):834--848, Apr. 2018.

\bibitem{chen2013knn}
Q.~Chen, D.~Li, and C.-K. Tang.
\newblock {KNN} matting.
\newblock {\em IEEE Transactions on Pattern Analysis and Machine Intelligence},
  35(9):2175--2188, 2013.

\bibitem{chen_counting_2017}
S.~W. Chen, S.~S. Shivakumar, S.~Dcunha, J.~Das, E.~Okon, C.~Qu, C.~J. Taylor,
  and V.~Kumar.
\newblock Counting {Apples} and {Oranges} {With} {Deep} {Learning}: {A}
  {Data}-{Driven} {Approach}.
\newblock {\em IEEE Robotics and Automation Letters}, 2(2):781--788, Apr. 2017.

\bibitem{chen2018blazingly}
Y.~Chen, J.~Pont-Tuset, A.~Montes, and L.~Van~Gool.
\newblock Blazingly fast video object segmentation with pixel-wise metric
  learning.
\newblock In {\em IEEE Conference on Computer Vision and Pattern Recognition},
  pages 1189--1198, 2018.

\bibitem{chuang2001bayesian}
Y.-Y. Chuang, B.~Curless, D.~H. Salesin, and R.~Szeliski.
\newblock A bayesian approach to digital matting.
\newblock In {\em IEEE Conference on Computer Vision and Pattern Recognition},
  pages 264--271. IEEE, 2001.

\bibitem{cordts2016cityscapes}
M.~Cordts, M.~Omran, S.~Ramos, T.~Rehfeld, M.~Enzweiler, R.~Benenson,
  U.~Franke, S.~Roth, and B.~Schiele.
\newblock The {Cityscapes} dataset for semantic urban scene understanding.
\newblock In {\em IEEE Conference on Computer Vision and Pattern Recognition},
  pages 3213--3223, 2016.

\bibitem{cremers_probabilistic_2007}
D.~Cremers, O.~Fluck, M.~Rousson, and S.~Aharon.
\newblock A probabilistic level set formulation for interactive organ
  segmentation.
\newblock In {\em Medical {Imaging} 2007: {Image} {Processing}}, volume 6512,
  page 65120V. International Society for Optics and Photonics, Mar. 2007.

\bibitem{cremers_review_2007}
D.~Cremers, M.~Rousson, and R.~Deriche.
\newblock A {Review} of {Statistical} {Approaches} to {Level} {Set}
  {Segmentation}: {Integrating} {Color}, {Texture}, {Motion} and {Shape}.
\newblock {\em International Journal of Computer Vision}, 72(2):195--215, Apr.
  2007.

\bibitem{deng2009imagenet}
J.~Deng, W.~Dong, R.~Socher, L.-J. Li, K.~Li, and L.~Fei-Fei.
\newblock Imagenet: A large-scale hierarchical image database.
\newblock In {\em IEEE Conference on Computer Vision and Pattern Recognition},
  pages 248--255. Ieee, 2009.

\bibitem{dias_semantic_2018}
P.~A. Dias and H.~Medeiros.
\newblock Semantic {Segmentation} {Refinement} by {Monte} {Carlo} {Region}
  {Growing} of {High} {Confidence} {Detections}.
\newblock In {\em Asian Conference on Computer Vision}, 2018.

\bibitem{dias_apple_2018}
P.~A. Dias, A.~Tabb, and H.~Medeiros.
\newblock Apple flower detection using deep convolutional networks.
\newblock {\em Computers in Industry}, 99:17--28, Aug. 2018.

\bibitem{Tabb2018Datasets}
P.~A. Dias, A.~Tabb, and H.~Medeiros.
\newblock Data from: {Multi}-species fruit flower detection using a refined
  semantic segmentation network, 2018.
\newblock http://dx.doi.org/10.15482/USDA.ADC/1423466.

\bibitem{dias_multispecies_2018}
P.~A. Dias, A.~Tabb, and H.~Medeiros.
\newblock Multispecies {Fruit} {Flower} {Detection} {Using} a {Refined}
  {Semantic} {Segmentation} {Network}.
\newblock {\em IEEE Robotics and Automation Letters}, 3(4):3003--3010, Oct.
  2018.

\bibitem{everingham_pascal_2015}
M.~Everingham, S.~M.~A. Eslami, L.~Van~Gool, C.~K.~I. Williams, J.~Winn, and
  A.~Zisserman.
\newblock The {Pascal} {Visual} {Object} {Classes} {Challenge}: {A}
  {Retrospective}.
\newblock {\em International Journal of Computer Vision}, 111(1):98--136, Jan.
  2015.

\bibitem{everingham_pascal_2010}
M.~Everingham, L.~Van~Gool, C.~K.~I. Williams, J.~Winn, and A.~Zisserman.
\newblock The {Pascal} {Visual} {Object} {Classes} ({VOC}) {Challenge}.
\newblock {\em International Journal of Computer Vision}, 88(2):303--338, June
  2010.

\bibitem{freedman_interactive_2005}
D.~Freedman and T.~Zhang.
\newblock Interactive graph cut based segmentation with shape priors.
\newblock In {\em {IEEE} {Conference} on {Computer} {Vision} and {Pattern}
  {Recognition}}, volume~1, pages 755--762 vol. 1, June 2005.

\bibitem{giuffrida_citizen_2018}
M.~V. Giuffrida, F.~Chen, H.~Scharr, and S.~A. Tsaftaris.
\newblock Citizen crowds and experts: observer variability in image-based plant
  phenotyping.
\newblock {\em Plant Methods}, 14:12, Feb. 2018.

\bibitem{he2018mask}
K.~He, G.~Gkioxari, P.~Doll{\'a}r, and R.~Girshick.
\newblock {Mask R-CNN}.
\newblock {\em {IEEE Transactions on Pattern Analysis and Machine
  Intelligence}}, 2018.

\bibitem{kawrykow2012phylo}
A.~Kawrykow, G.~Roumanis, A.~Kam, D.~Kwak, C.~Leung, C.~Wu, E.~Zarour,
  L.~Sarmenta, M.~Blanchette, J.~Waldisp{\"u}hl, et~al.
\newblock Phylo: a citizen science approach for improving multiple sequence
  alignment.
\newblock {\em PloS one}, 7(3):e31362, 2012.

\bibitem{kolesnikov_seed_2016}
A.~Kolesnikov and C.~H. Lampert.
\newblock Seed, expand and constrain: Three principles for weakly-supervised
  image segmentation.
\newblock In {\em European Conference on Computer Vision}, pages 695--711.
  Springer, 2016.

\bibitem{krahenbuhl2012crf}
P.~Kr{\"{a}}henb{\"{u}}hl and V.~Koltun.
\newblock {Efficient Inference in Fully Connected CRFs with Gaussian Edge
  Potentials}.
\newblock In {\em Advances in Neural Information Processing Systems}, pages
  109--117, 2011.

\bibitem{Li2016FCIS}
Y.~Li, H.~Qi, J.~Dai, X.~Ji, and Y.~Wei.
\newblock {Fully Convolutional Instance-aware Semantic Segmentation}.
\newblock In {\em IEEE Conference on Computer Vision and Pattern Recognition},
  pages 2359--2367, 2017.

\bibitem{lin_scribblesup:_2016}
D.~Lin, J.~Dai, J.~Jia, K.~He, and J.~Sun.
\newblock {ScribbleSup}: {Scribble}-{Supervised} {Convolutional} {Networks} for
  {Semantic} {Segmentation}.
\newblock In {\em {IEEE} {Conference} on {Computer} {Vision} and {Pattern}
  {Recognition}}, pages 3159--3167, June 2016.

\bibitem{Lin2014coco}
T.-Y. Lin, M.~Maire, S.~Belongie, J.~Hays, P.~Perona, D.~Ramanan,
  P.~Doll{\'a}r, and C.~L. Zitnick.
\newblock {Microsoft COCO: Common objects in context}.
\newblock In {\em {European Conference on Computer Vision}}, pages 740--755.
  Springer, 2014.

\bibitem{liu_interactive_2012}
Y.~Liu and Y.~Yu.
\newblock Interactive {Image} {Segmentation} {Based} on {Level} {Sets} of
  {Probabilities}.
\newblock {\em IEEE Transactions on Visualization and Computer Graphics},
  18(2):202--213, Feb. 2012.

\bibitem{mahadevan2018iter}
S.~Mahadevan, P.~Voigtlaender, and B.~Leibe.
\newblock Iteratively trained interactive segmentation.
\newblock In {\em British Machine Vision Conference}, 2018.

\bibitem{dextr2018}
K.~Maninis, S.~Caelles, J.~Pont-Tuset, and L.~{Van Gool}.
\newblock Deep extreme cut: From extreme points to object segmentation.
\newblock In {\em IEEE Conference on Computer Vision and Pattern Recognition},
  2018.

\bibitem{mortensen1995intelligent}
E.~N. Mortensen and W.~A. Barrett.
\newblock Intelligent scissors for image composition.
\newblock In {\em Computer Graphics and Interactive Techniques}, pages
  191--198. ACM, 1995.

\bibitem{rajchl_deepcut:_2017}
M.~Rajchl, M.~C.~H. Lee, O.~Oktay, K.~Kamnitsas, J.~Passerat-Palmbach, W.~Bai,
  M.~Damodaram, M.~A. Rutherford, J.~V. Hajnal, B.~Kainz, and D.~Rueckert.
\newblock {DeepCut}: {Object} {Segmentation} {From} {Bounding} {Box}
  {Annotations} {Using} {Convolutional} {Neural} {Networks}.
\newblock {\em IEEE Transactions on Medical Imaging}, 36(2):674--683, Feb.
  2017.

\bibitem{rother_grabcut:_2004}
C.~Rother, V.~Kolmogorov, and A.~Blake.
\newblock "{GrabCut}": {Interactive} {Foreground} {Extraction} {Using}
  {Iterated} {Graph} {Cuts}.
\newblock In {\em {ACM transactions on graphics (TOG)}}, volume~23, pages
  309--314. ACM, 2004.

\bibitem{russell_labelme:_2008}
B.~C. Russell, A.~Torralba, K.~P. Murphy, and W.~T. Freeman.
\newblock {LabelMe}: {A} {Database} and {Web}-{Based} {Tool} for {Image}
  {Annotation}.
\newblock {\em International Journal of Computer Vision}, 77(1-3):157--173, May
  2008.

\bibitem{schwartz2003paradox}
B.~Schwartz.
\newblock {\em The Paradox of Choice: Why More Is Less}.
\newblock Harper Perennial. HarperCollins, 2003.

\bibitem{stutz2017sppx}
D.~Stutz, A.~Hermans, and B.~Leibe.
\newblock Superpixels: An evaluation of the state-of-the-art.
\newblock {\em Computer Vision and Image Understanding}, 2017.

\bibitem{tang_normalized_2018}
M.~Tang, A.~Djelouah, F.~Perazzi, Y.~Boykov, and C.~Schroers.
\newblock Normalized {Cut} {Loss} for {Weakly}-{Supervised} {CNN}
  {Segmentation}.
\newblock {\em IEEE Conference on Computer Vision and Pattern Recognition},
  IEEE Conference on Computer Vision and Pattern Recognition:10, 2018.

\bibitem{tangseng_looking_2017}
P.~Tangseng, Z.~Wu, and K.~Yamaguchi.
\newblock Looking at {Outfit} to {Parse} {Clothing}.
\newblock {\em arXiv:1703.01386 [cs]}, Mar. 2017.
\newblock arXiv: 1703.01386.

\bibitem{von2004labeling}
L.~Von~Ahn and L.~Dabbish.
\newblock Labeling images with a computer game.
\newblock In {\em SIGCHI conference on Human factors in computing systems},
  pages 319--326. ACM, 2004.

\bibitem{von2008designing}
L.~Von~Ahn and L.~Dabbish.
\newblock Designing games with a purpose.
\newblock {\em Communications of the ACM}, 51(8):58--67, 2008.

\bibitem{von2006verbosity}
L.~Von~Ahn, M.~Kedia, and M.~Blum.
\newblock Verbosity: a game for collecting common-sense facts.
\newblock In {\em SIGCHI conference on Human Factors in computing systems},
  pages 75--78. ACM, 2006.

\bibitem{von2006peekaboom}
L.~Von~Ahn, R.~Liu, and M.~Blum.
\newblock Peekaboom: a game for locating objects in images.
\newblock In {\em SIGCHI conference on Human Factors in computing systems},
  pages 55--64. ACM, 2006.

\bibitem{vondrick_efficiently_2013}
C.~Vondrick, D.~Patterson, and D.~Ramanan.
\newblock Efficiently {Scaling} up {Crowdsourced} {Video} {Annotation}: {A}
  {Set} of {Best} {Practices} for {High} {Quality}, {Economical} {Video}
  {Labeling}.
\newblock {\em International Journal of Computer Vision}, 101(1):184--204, Jan.
  2013.

\bibitem{wang2014touchcut}
T.~Wang, B.~Han, and J.~Collomosse.
\newblock Touchcut: Fast image and video segmentation using single-touch
  interaction.
\newblock {\em Computer Vision and Image Understanding}, 120:14--30, 2014.

\bibitem{zhu2015targeting}
Q.~Zhu, L.~Shao, X.~Li, and L.~Wang.
\newblock Targeting accurate object extraction from an image: A comprehensive
  study of natural image matting.
\newblock {\em IEEE Transactions on Neural networks and Learning Systems},
  26(2):185--207, 2015.

\end{thebibliography}
}

\end{document}